\documentclass{article}
\usepackage{arxiv}

\usepackage[utf8]{inputenc} 
\usepackage[T1]{fontenc}    
\usepackage{hyperref}       
\usepackage{url}            
\usepackage{booktabs}       
\usepackage{amsfonts}       
\usepackage{nicefrac}       
\usepackage{microtype}      
\usepackage{lipsum}		
\usepackage{graphicx}
\usepackage{natbib}
\usepackage{doi}
\usepackage[table]{xcolor}
\usepackage{amsmath}
\usepackage{amssymb}
\usepackage{amsthm}
\usepackage{multirow}
\usepackage{titlesec}

\pdfoutput=1
\usepackage[pdftex]{}

\newcommand{\mypar}[1]{\vspace{3pt}\noindent\textbf{#1~}}
\newcommand{\zz}{\mathbf{z}}
\newcommand{\xx}{\mathbf{x}}

\newcommand{\mr}[1]{\mathrm{#1}}
\newcommand{\Loss}{\mathcal{L}}
\newcommand{\ppm}{\,\scriptsize$\pm$}
\newcommand{\dorowcolors}{\rowcolors{2}{gray!15}{white}}

\newtheorem{theorem}{Theorem}[section]
\newtheorem{lemma}[theorem]{Lemma}

\title{ClusT3: Information Invariant Test-Time Training}

\date{} 					

\author{Gustavo A. Vargas Hakim\thanks{Equal contribution} \And David Osowiechi\footnotemark[1] \And Mehrdad Noori \And Milad Cheraghalikhani \And Ismail Ben Ayed \And Christian Desrosiers}
\date{LIVIA, ÉTS Montréal, Canada \\ International Laboratory on Learning Systems (ILLS), \\ McGILL - ETS - MILA - CNRS - Université Paris-Saclay - CentraleSupélec, Canada \texttt{gustavo-adolfo.vargas-hakim.1@ens.etsmtl.ca, david.osowiechi.1@ens.etsmtl.ca,
mehrdad.noori.1@ens.etsmtl.ca, milad.cheraghalikhani.1@ens.etsmtl.ca\\
ismail.benayed@etsmtl.ca,  christian.desrosiers@etsmtl.ca}}

\renewcommand{\shorttitle}{ClusT3: Information Invariant Test-Time Training}

\hypersetup{
pdftitle={ClusT3},
pdfsubject={cs.ML, cs.AI, cs.CV},
pdfauthor={Gustavo A. Vargas Hakim, David Osowiechi, Mehrdad Noori, Milad Cheraghalikhani, Ismail Ben Ayed, Christian Desrosiers},
pdfkeywords={Test-time Adaptation, Image Classification, Mutual Information, Unsupervised Training},
}

\begin{document}
\maketitle

\begin{abstract}
   Deep Learning models have shown remarkable performance in a broad range of vision tasks. However, they are often vulnerable against domain shifts at test-time. Test-time training (TTT) methods have been developed in an attempt to mitigate these vulnerabilities, where a secondary task is solved at training time simultaneously with the main task, to be later used as an self-supervised proxy task at test-time. In this work, we propose a novel unsupervised TTT technique based on the maximization of Mutual Information  between multi-scale feature maps and a discrete latent representation, which can be integrated to the standard training as an auxiliary clustering task. Experimental results demonstrate competitive classification performance on different popular test-time adaptation benchmarks.
\end{abstract}

\section{Introduction}
\label{sec:intro}

The domain invariance hypothesis has been key to the success of deep learning methods for computer vision. In this hypothesis, the training and testing data are both assumed to be drawn from the same distribution, which rarely holds in practical settings. Moreover, it has been shown in numerous studies that the performance in classification and segmentation can drop significantly when domain shifts are present \cite{Recht2018,visda}. In response, Domain Adaptation (DA) studies the adaptation of learning algorithms to new domains, when different types of domain shifts are present in the test data. From this field, two promising directions have emerged: Domain Generalization and Test-Time Adaptation. On the one hand, Domain Generalization (DG) \cite{dg1,dg2,dg3,dg4,dgsurvey} assumes a model is trained on a large source dataset composed of different domains, and evaluates the performances on new domains at test-time. On the other hand, Test-Time Adaptation (TTA) \cite{tent2021,shot2021,sita2021,lame2022}   adapts the model to test data \emph{on the fly}, typically adjusting to subsets of the new domain (e.g., mini-batches) each time. In TTA, there is no supervision from the testing samples nor access to the source domain, which makes it a challenging, yet realistic problem. The main limitation of DG is the requirement of a large amount of training data from different domains, without the guarantee that the model generalizes well to the (virtually unlimited) possible new domains it may encounter. TTA methods do not have this issue. However, they are highly sensitive to the choice of the unsupervised loss functions deployed at test-time, which may severely hurt the performances. 

\begin{figure*}[ht!]
    \centering    \begin{small}\setlength{\tabcolsep}{8pt}
    \begin{tabular}{ccc}    
\includegraphics[width=0.295\linewidth]{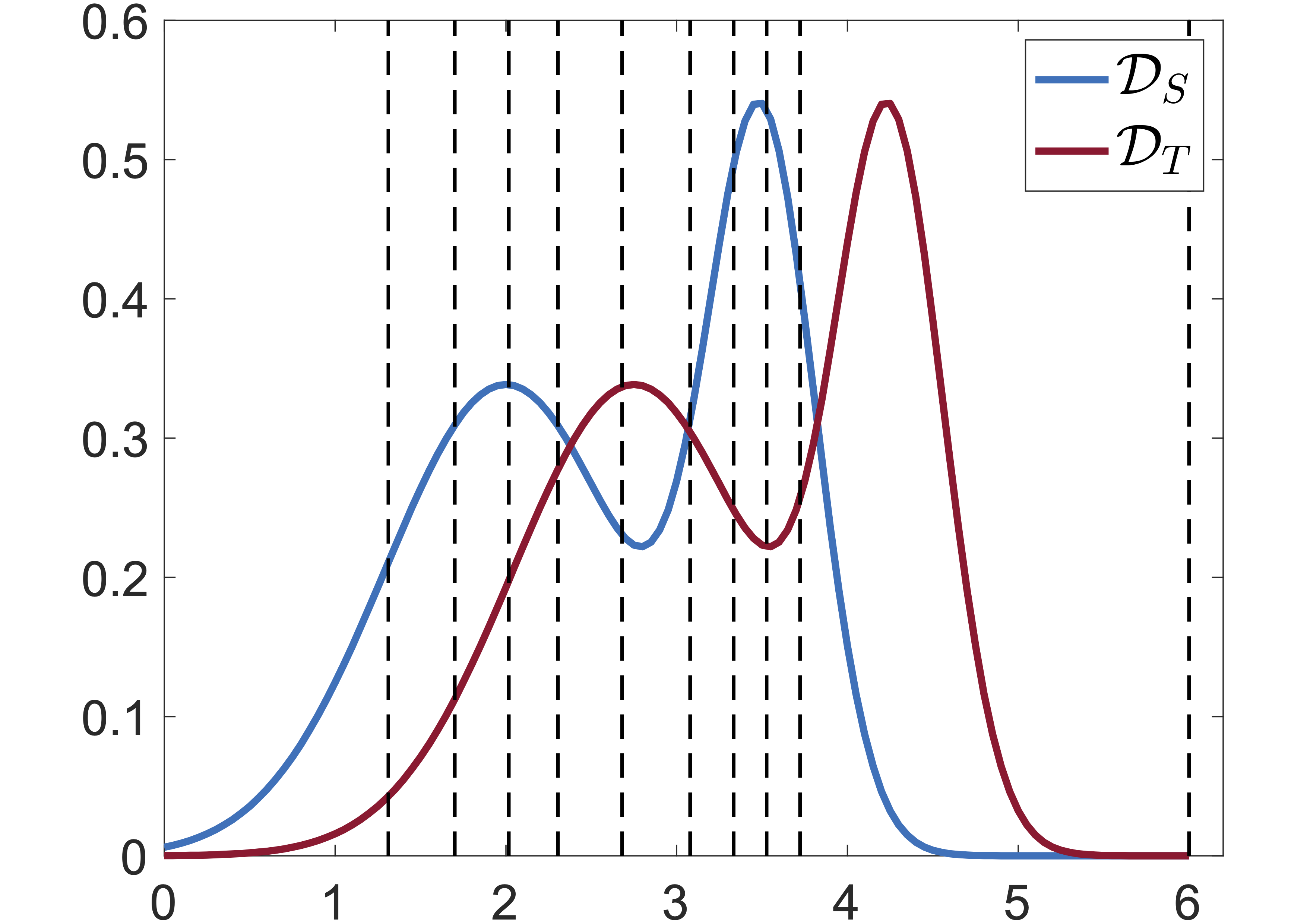} &     
    \includegraphics[width=0.295\linewidth]{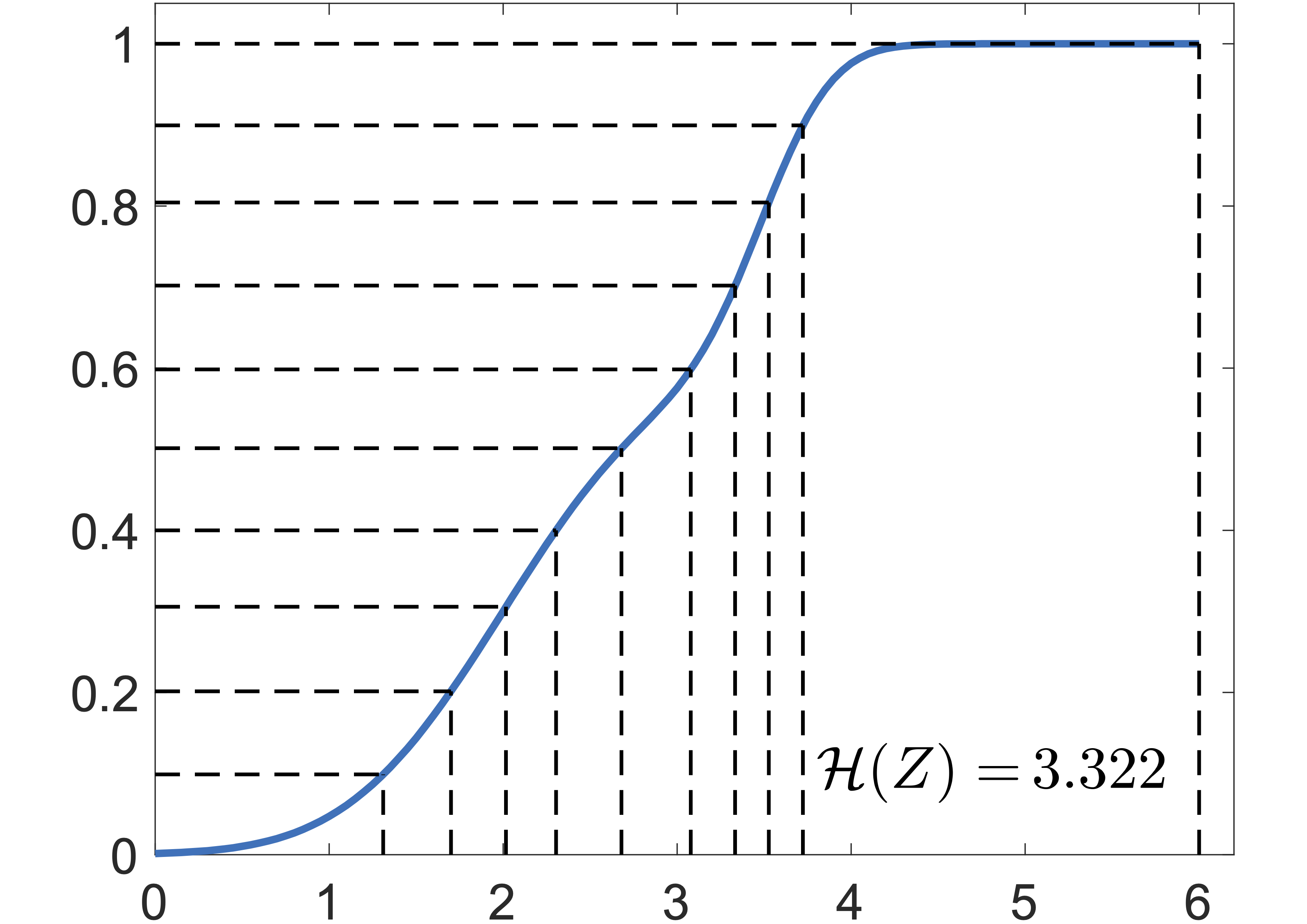}
    &     
    \includegraphics[width=0.295\linewidth]{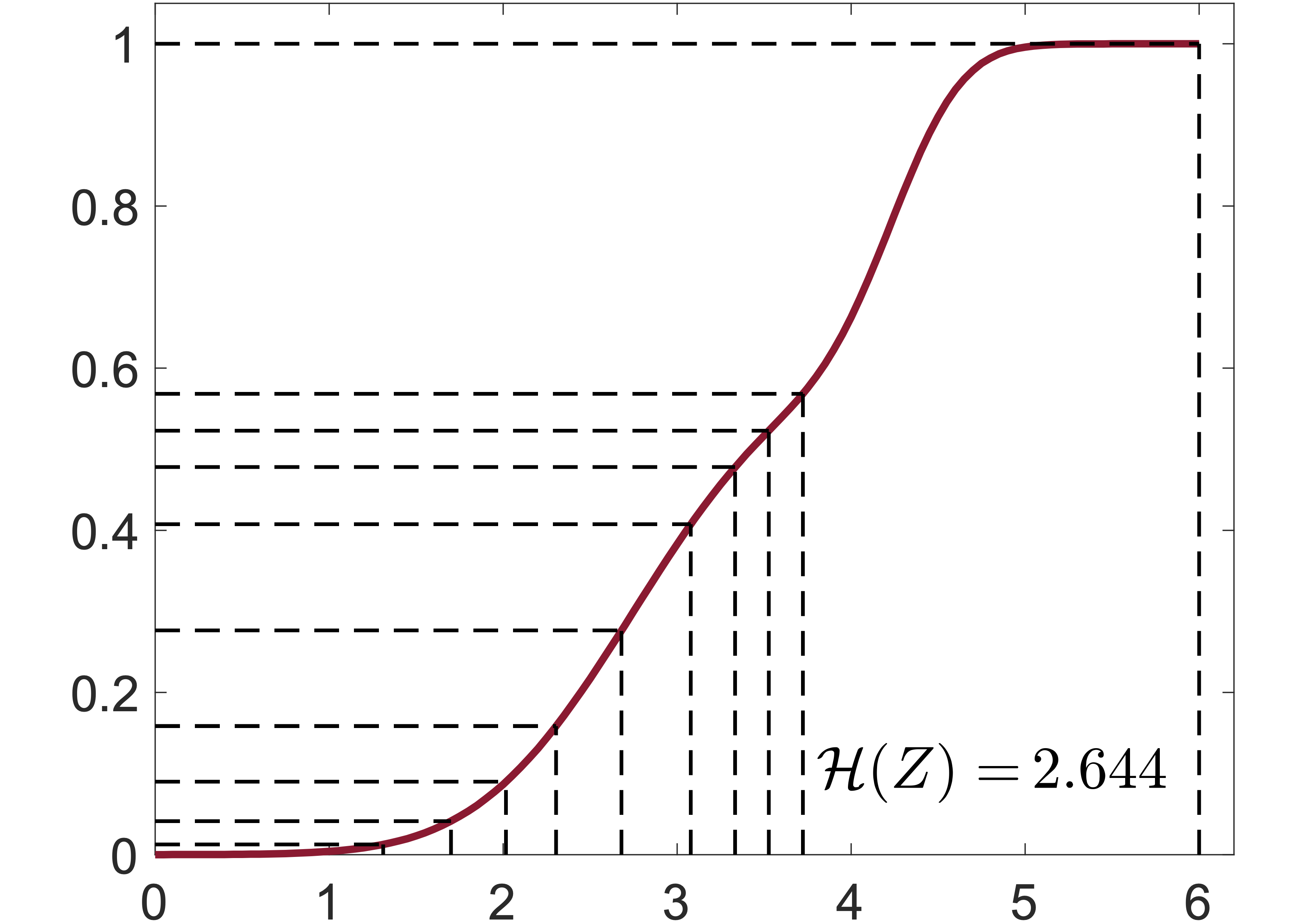}
    \\
    (a) Source $\mathcal{D}_{S}$ and Target $\mathcal{D}_{T}$ distributions & 
    (b) Cumulative density function of $\mathcal{D}_{S}$ &
   (c) Cumulative density function of $\mathcal{D}_{T}$\\[4pt]
    \end{tabular}
    \end{small}
    \caption{\textbf{Illustration of our Information Invariant TTT method on a 1D feature space}. (\emph{a}) The clustering of source features $\mathbf{x}$ (\emph{blue}) into $K\!=\!10$ regions, maximizing the entropy of the cluster marginal distribution $\mathcal{H}(Z)$, is such that regions have the same probability mass in the source distribution. At test-time, the probability density function of the target domain (\emph{red}) is shifted, which results in a different clustering of features. (\emph{b}) The optimal clustering corresponds to dividing the cumulative density function (CDF) in even steps, giving a cluster marginal entropy of $\mathcal{H}(Z)\!=\!\log_2(K)\!\approx\!3.332$. (\emph{c})  Since the CDF of the target is not divided in even steps, the mutual information between features $\mathbf{x}$ and clusters $\mathbf{z}$ is no longer maximized. Note: we assume that cluster assignments are confident, i.e., $\mathcal{H}(Z|X)\!\approx\!0$ and thus $\mathcal{I}(Z;X)=\mathcal{H}(Z)\!-\!\mathcal{H}(Z|X)\approx\mathcal{H}(Z)$.}
    \label{fig:im}
\end{figure*}

Test-Time Training (TTT) \cite{ttt,ttt++,tttmask,tttflow} is an attractive variant of TTA, where an auxiliary task is learned from the training data (source domain) and later used at test-time to update a model. Typically, unsupervised and self-supervised tasks are chosen, as they allow for an adaptation process that does not require any label. Moreover, the joint, two-task training protocol for the source domain provides \emph{momentum} at test-time, enabling the use of a loss function that is not completely foreign to the model.

Inspired by the recent success of Mutual-Information (MI) maximization in several learning tasks, such as representation learning \cite{ji2019invariant,hu2017learning,oord2018representation,tschannenmutual}, deep clustering \cite{Jabi21} and few-shot learning \cite{TIM}, we propose an information invariant TTT method called ClusT3. Our method maximizes the MI between the feature maps at different scales and discrete latent representations related to clustering. The main idea is that the amount of information between the features and their corresponding discrete encoding should remain constant in both the source and target domains (see Fig.~\ref{fig:im}). Toward this goal, we introduce an auxiliary task that performs information-maximization clustering while training on the source examples. At test time, we use the MI between the features and cluster assignments as a measure of representation quality, and maximize the MI as objective for test-time adaptation. Unlike previous TTT approaches, which rely on problem-specific, self-supervised learning strategies, our auxiliary clustering task is problem-agnostic and could be added on top of any model via a low-dimensional linear projection. Test-time adaptation could also be done using only the test samples, without any type of distilled information from the source domain. On the technical side, minimal architectural changes are needed, and the joint training approach is more efficient than proceeding with multiple, complex and time-consuming steps. 

Our contributions could be summarized as follows:
\begin{itemize}
    \item We propose a novel Test-Time Training approach based on maximizing the MI between feature maps and discrete representations learned in training. At test time, adaptation is achieved based on the principle that information between the features and their discrete representation should remain constant across domains. 
    \item ClusT3 is evaluated across a series of challenging TTA scenarios, with different types of domain shifts, obtaining competitive performance compared to previous methods. 
    \item To the best of our knowledge, this is the first Unsupervised Test-Time Training approach using  a joint training based on the MI and linear projectors. Our approach is lightweight and more general than its previous self-supervised counterparts. 
\end{itemize}
The rest of this paper is organized as follows. Section~\ref{sec:previous} presents previous work in both TTA and TTT. Section~\ref{sec:method} introduces the ClusT3 method with the experimental setting to evaluate it in Section~\ref{sec:experiments}. Experimental results and discussions are provided in Section~\ref{sec:results}, and the closing conclusions are given in Section~\ref{sec:conclusion}.

\section{Related Work}
\label{sec:previous}

\mypar{Test-Time Adaptation.} The goal of TTA is to adapt a pre-trained model to a target dataset \emph{on the fly}, i.e., as batches of data appear. Additional challenges include (1) the inaccessibility of source samples, which makes direct domain alignment impossible, (2) the lack of label supervision, which makes using unsupervised losses necessary, and (3) the fact that there is no access to all the target distribution, as the data come in the form of batches and not as a whole dataset. Adaptation can then be performed on different components of a network, such as the feature extractor, the classifier, or even the whole network. 

Prediction Time Batch Normalization (PTBN) \cite{PTBN} proposes to use the feature mean and variance from the batch of test samples as statistics in the batch norm layers. TENT \cite{tent2021} instead focuses its adaptation on the affine parameters of the batch normalization layers only, based on the conditional entropy loss of the predictions. By updating linear parameters, the model can be more easily optimized and the source knowledge is preserved. SHOT \cite{shot2021} also freezes the classifier, but adapts the entire feature encoder by minimizing the uncertainty of predictions (low conditional entropy) while making them class-balanced (high entropy of class marginals). To circumvent the problem of erroneous predictions, the model also uses a pseudo-labeling mechanism coupled with cross-entropy as part of the final loss. LAME \cite{lame2022} reduces the adaptation focus even more, by only refining the classifier's predictions on target batches. In a spirit similar to that of SHOT, LAME utilizes a KL divergence loss on the class marginal distribution to make it more uniform, and a feature-level Laplacian regularizer to encourage concise clustering based on similarity. Test-time adaptation is performed using a closed-form iterative optimization process. 

\mypar{Test-Time Training.} In line with TTA methods, TTT seeks to update a model at test-time using an auxiliary task that has been trained along the main classification objective during source training. TTT \cite{ttt}, which is among the first of such techniques, uses a Y-shaped architecture where a self-supervised rotation prediction network is attached to an arbitrary layer in the feature extractor of a CNN. A standard supervised cross-entropy loss ($ \Loss_{\mr{CE}}$) is optimized jointly with the auxiliary self-supervised loss $\Loss_{\mr{aux}}$ of the secondary branch, as follows:
\begin{equation}
    \Loss_{\mr{TTT}} = \Loss_{\mr{CE}} + \lambda \Loss_{\mr{aux}}
    \label{eq:tttloss}
\end{equation}
At test-time, only the layers connected to the secondary branch are updated. The loss in Eq.~(\ref{eq:tttloss}) served as basis for subsequent TTT methods. TTT++ \cite{ttt++} introduced contrastive learning as the secondary task, similarly to TTT. However, to further improve performance at test-time, the statistics of source data are computed from a preserved queue of source feature maps. These statistics are then used for alignment with target data, thus regularizing the contrastive loss. TTT-MAE \cite{tttmask} proposes using Masked Autoencoders (MAE) \cite{mae} as the second branch for test-time training. This approach also introduced Vision Transformers \cite{vit} in the context of TTA and TTT. Different from standard TTT methods, TTTFlow \cite{tttflow} first pre-trains the model with a standard cross-entropy loss and then adds a Normalizing Flow (NF) \cite{realnvp,glow} as a secondary task on top of early encoder layers. The NF is trained on source data independently of the classification task, by maximizing the log likelihood of source examples mapped to a simple distribution (Gaussian). The same loss function is later used to adapt the feature extractor for target data.



\begin{figure*}[ht!]
    \centering
\includegraphics[width=.65\linewidth]
{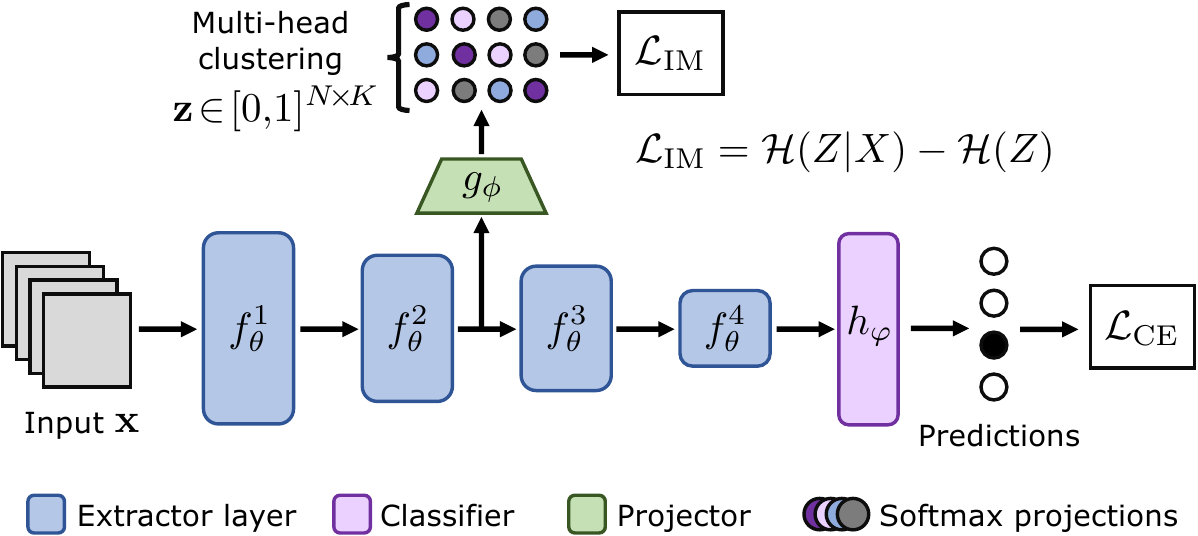}
    \caption{The configuration of ClusT3. A projector $g_\phi$ is plugged to the output of a feature extractor layer block to compute a set of $N$, $K-$dimensional latent points $\zz$ that are clustered through Information Maximization ($\mathcal{L}_{\mr{IM}}$). The cross-entropy loss ($\mathcal{L}_{\mr{CE}}$) is used for the classification component of training.}
    \label{fig:clust3}
\end{figure*}

\section{Method}
\label{sec:method}

In this section, we present a formal definition of Test-Time Training, followed by the description of our ClusT3 method.

\subsection{Problem formulation}

Let $P(\mathcal{X}_{s},\mathcal{Y}_{s})$ be the joint distribution that represents the source domain, where $\mathcal{X}_{s}$ and $\mathcal{Y}_{s}$ are the input and label spaces, respectively. Similarly, $P(\mathcal{X}_{t},\mathcal{Y}_{t})$ corresponds to the target domain distribution, with inputs and labels $\mathcal{X}_{t}$ and $\mathcal{Y}_{t}$. In this work, we consider a likelihood shift \cite{lame2022} between the source and target datasets, i.e., $P(\mathcal{X}_{s}|\mathcal{Y}_{s}) \neq P(\mathcal{X}_{t}|\mathcal{Y}_{t})$, with both domain sharing the same label space ($\mathcal{Y}_{s} = \mathcal{Y}_{t}$).

A standard TTT-based model is composed of a feature extractor $f_{\theta}$, a classifier $h_{\varphi}$, and an auxiliary module $g_{\phi}$, all collected inside the functional $F(f_{\theta}, h_{\varphi},g_{\phi})$. During training, the goal is to learn $F_{s}: \mathcal{X}_{s} \rightarrow \mathcal{Y}_{s}$ using Eq.~(\ref{eq:tttloss}), where the unsupervised loss $\mathcal{L}_{\mr{aux}}$ is chosen to be related to the auxiliary task $g_{\phi}$. At test-time, only the unsupervised loss is used to adapt the model, such that we learn an adapted function $F_{t}: \mathcal{X}_{t} \rightarrow \mathcal{Y}_{t}$.

\subsection{Proposed method}
\label{subseq:method}

ClusT3 is built on the formulation of previous work on TTT, following Eq.~(\ref{eq:tttloss}) and using modules plugged to the feature extractor. As shown in Fig.~\ref{fig:clust3}, we learn a discretized encoding of feature maps in the encoder using a clustering strategy based on MI maximization. Denote as $f_{\theta}(\xx)\!\in\!\mathbb{R}^{N \times C}$ the combined features of examples in a batch of size $B$, where the first dimension $N\!=\!B\!\cdot\!W\!\cdot\!H$ is obtained by flattening along the batch index and feature map dimensions. We use a shallow projector $g_{\phi}$ to map $f_{\theta}(\xx)$ into a set of $K$-cluster probability distributions $\zz = g_{\phi}(f_{\theta}(\xx))\!\in\![0,1]^{N\!\times\!K}$. In its simplest form, this projector is implemented by a single linear mapping followed by a softmax. A more complex projector, comprised of additional linear layers with ReLU activation can also be employed. We train the projector by maximizing the MI between $\xx$ and its discrete representation $\zz$:
\begin{align}
\label{eq:im}
\mathcal{L}_{\mr{IM}} & \, = \, -\mathcal{I}(X; Z) \, = \, 
\mathcal{H}(Z|X) - \mathcal{H}(Z) \\
 & \, = \, -\frac{1}{N}\sum_{i=1}^{N}\sum_{K\!=\!1}^{K}z_{ik}\log{z_{ik}} + \sum_{K\!=\!1}^{K}\overline{z}_{k}\log\overline{z}_{k}\nonumber
\end{align}
where $\overline{z}_k\!=\!\frac{1}{N}\sum_i z_{ik}$ is the average probability of cluster $K$. The first term, $\mathcal{H}(\zz|\xx)$, is the conditional entropy of $\zz$ given $\xx$. Minimizing this term enforces the model to make confident assignments of examples to clusters. On the other hand, the term $\mathcal{H}(\zz)$ corresponds to the entropy of the cluster marginal distribution. Maximizing this term encourages the clusters to be balanced, and avoids the trivial solution of mapping all examples to a single cluster. 

In connection to information theory, our approach seeks a compressed encoding $Z$ of features $U\!=\!f(X)$, modeled by a Markov chain $X\!\to\!U\!\to\!Z$, which best preserves information. Following the data processing inequality, we necessarily have that $\mathcal{I}(X; U) \geq \mathcal{I}(X; Z)$. The clustering defined by random variable $Z$ divides the feature space in $K$ regions. To maximize MI, it is known that the clustering must satisfy two conditions. First, it should divide the feature space in regions $\{\mathcal{R}_k\}_{k\!=\!1}^K$ of equal probability mass, i.e., $\int_{\mathcal{R}_k} p(u) du = \int_{\mathcal{R}_{k'}} p(u) du$, for any $k,k'$
\cite{mackay2003information}. Second, the features falling into each region $\mathcal{R}_k$ should be similar, i.e., the entropy of $U$ given $Z$ should be low. Accordingly, increasing the number $K$ of clusters leads to a higher MI. Assuming that the clustering in $Z$ is a good representation of the distribution of features $U$, a shift in this distribution at test-time is likely to decrease MI since the shifted distribution is not well represented by $Z$.  

\mypar{Multi-scale clustering.} In ClusT3, different projectors can be independently placed on top of different layer blocks of a CNN (e.g., ResNet). In such case, the output of the $\ell$-th layer is now written as $\zz^{\ell}\!=\!g_{\phi}(f_{\theta}^{\ell}(\xx))$. At training time, the model learns with a combined loss 
\begin{equation}
    \mathcal{L}_{\mr{CT3}} \, = \, \mathcal{L}_{\mr{CE}} + \sum_{\ell=j}^{J}\mathcal{L}_{\mr{IM}}^{\ell}
\label{eq:loss}
\end{equation}
where $j$ the index of the layer from which the first projector is connected. At test-time, the classifier $h_\varphi$ and the projectors $\{g_{\phi}^{\ell}\}_{\ell=j}^{J}$ are frozen, and only the feature extractor $f_{\theta}$ up to layer $J$ is updated based on the IM loss of Eq.~(\ref{eq:im}). It is worth noting that the gradient flow is going to affect only the layer blocks connected to the projectors and the ones before. The hypothesis is that the latent space of the feature maps should be information invariant across domains, thus updating the encoder to maintain a high mutual information should also improve classification accuracy.

\mypar{Multi-head clustering.} As mentioned above, an encoding that better preserves information can be achieved by using a larger number of clusters. In practice, doing so might give poor results since the constraint of having balanced clusters (low entropy of the marginal) then becomes too restrictive. As better alternative, we propose a multi-head clustering strategy where multiple projectors $\{g_{\phi}^{\ell,c}\}_{c=1}^{C}$ are trained for a given layer $\ell$ and the loss $\mathcal{L}_{\mr{IM}}^\ell$ for that layer is the sum of MI losses for all its projectors. The following lemma relates this strategy to our previous information theory analysis.

\begin{lemma}
Let $\mathcal{Z} = \{Z_1, \ldots, Z_C\}$ be a set of random discrete variables representing $C$ cluster assignments of features $X$. The MI between $X$ and $\mathcal{Z}$ is bounded as follows
\begin{equation}
\max_c \, \mathcal{H}(Z_c) - \sum_c \mathcal{H}(Z_c|X) \, \leq \, \mathcal{I}(X; \mathcal{Z}) \, \leq \sum_c \mathcal{I}(X; Z_c)\nonumber
\end{equation}
\begin{proof}
We start by writing the MI between $X$ and $\mathcal{Z}$ as
\begin{equation}
\mathcal{I}(X; \mathcal{Z}) \, = \, \mathcal{H}(Z_1, \ldots, Z_C) - \mathcal{H}(Z_1, \ldots, Z_C | X)\nonumber.
\end{equation}
The second term on the right simplifies as
\begin{align}
\mathcal{H}(Z_1, \ldots, Z_C | X) & \, = \, -\mathbb{E}\big[\, \log p(Z_1, \ldots, Z_C | X) \,\big] \nonumber\\
& \, = \, -\mathbb{E}\big[ \, \sum_c \log p(Z_c | X) \, \big] \nonumber\\
& \, = \, \sum_c \mathcal{H}(Z_c | X)\nonumber
\end{align}
where we used the fact that the $Z_c$ variables are conditionally independent given $X$. To complete the proof, we use the following two properties of entropy: $\mathcal{H}(Z_1,\ldots,Z_C) \leq \sum_c \mathcal{H}(Z_c)$ and $\mathcal{H}(Z_1,\ldots,Z_C) \geq \max_c \mathcal{H}(Z_c)$.
\end{proof}
\end{lemma}
Note that the upper bound on $\mathcal{I}(X; \mathcal{Z})$, which corresponds to our multi-head clustering objective, is tight if the clustering variables $Z_c$ are statistically independent. Although we do not enforce this constraint, since our objective maximizes $\mathcal{H}(Z_c)$ for \emph{each} cluster, the lower bound of the lemma tells us that we can indirectly maximize mutual information with the same objective.    

\section{Experimental Setup}
\label{sec:experiments}

ClusT3 is evaluated on four popular TTA/TTT benchmarks, comprehending different types of domain shifts.  The first two benchmarks are based on the CIFAR-10 dataset \cite{cifar10} as the source domain. It contains 50,000 images from 10 different categories. 

\mypar{Common image corruptions.} First, we study adaptation on the CIFAR-10-C \cite{cifar10c} dataset, which consists of 15 different corruption types (e.g., Gaussian noise, frost, etc.) with 10,000 images, 10 classes, and 5 different severity levels for each type. This results in 75 evaluation scenarios. We then extend the evaluation to CIFAR-100-C, scaling the number of classes to 100.

\mypar{Natural domain shift.} We also evaluate the performance of our method in a natural domain shift setting, i.e., classifying images that were manually selected to diverge from those seen in training.  The CIFAR-10.1 dataset \cite{cifar10.1} is used for this experiment, consisting of 2,000 images strategically sampled from CIFAR-10 to highly differ from training data.

\mypar{Sim-to-real domain shift.} ClusT3 is finally assessed in the context of large-scale adaptation from simulation to real images. The VisDa-C dataset \cite{visda} offers a benchmark with a source dataset based on 3D renderings of 12 different object categories, accumulating a total of 152,397 images. The test set comprises 72,372 video frames, corresponding to real images of the same classes.

\subsection{Joint training}

For the joint training on the CIFAR-10 dataset \cite{cifar10}, we followed previous research and trained our model for 350 epochs with SGD, using a batch size of 128 images and an initial learning rate of 0.1 which is reduced by a factor of 10 at epochs 150 and 250. For VisDA-C, the model is warm-started with pre-trained weights from ImageNet \cite{imagenet}, according to the protocol in \cite{tent2021,ttt++,shot2021}, and then trained for 100 epochs with a batch size of 100, using SGD with a learning rate of 0.001. The training was executed on four 16 GB NVIDIA V100 GPUs.

\subsection{Test-time adaptation}

At test-time, projectors are used to detect distribution shift with the IM loss. For all the experiments with CIFAR-10-C and CIFAR-10.1, we keep a batch size of 128, and use the ADAM optimizer with $10^{-5}$ as learning rate. For VisDA-C, we used a batch size of 32 images with the same aforementioned learning rate.
We update the extractor and the statistics of all the BatchNorm layers.
To avoid the error accumulation associated to optimization, we reset our weights to the initial source ones after  adapting to each batch.
This way, each batch can have different corruptions as assumed by \cite{ttt} in their offline mode. Our codebase can be found in \url{https://github.com/dosowiechi/ClusT3.git}.

\section{Results and discussion}
\label{sec:results}

First, we perform a series of ablation experiments on the CIFAR-10-C dataset, and then compare ClusT3 against state-of-art approaches. Afterward, we extend our evaluation to natural domain shift using the CIFAR-10.1 dataset and sim-to-real domain shift with the VisDA-C dataset. For all methods, we compute the accuracy for 1, 3, 5, 10, 20, 50 and 100 iterations and report the maximum accuracy when we experiment on CIFAR-10-C and CIFAR-10.1 and do the same for VisDAC by adapting for 1, 3, 10, 15, and 20 iterations. For all experiments, we report the mean and standard deviation accuracy obtained over 3 runs with different random seeds.

\subsection{Object recognition on corrupted images}

First, we evaluate ClusT3 on the CIFAR-10-C dataset across the 15 different corruptions. For the following experiments, we focus solely on the Level 5, as it is the most challenging adaptation scenario. Extensive results on all the severity levels can be found in the supplementary material.

\mypar{On which layers should projectors be placed?} We compare the accuracy of ClusT3 on different combinations of projectors. The goal is to determine which layers are the most useful to adapt at test-time. 
In Table~\ref{tab:LayersComparison}, the results show that only taking the first two encoder layers provides more effective results. Indeed, as assumed in \cite{ttt,ttt++,tttflow}, the first layers seem to contain the most important domain-related information. This finding also aligns with empirical evidence demonstrating that different layers are sensitive to different types of domain shifts \cite{surgical}. Hence, in subsequent experiments, we keep projectors on Layer 1 and Layer 2. 

\begin{table}[!t]
    \centering
    \begin{small}
    \dorowcolors
    \begin{tabular}{l|ccc}
    \toprule
        ~ & Gaussian Noise & Shot Noise & Snow \\ \midrule
        Layer 1 & 70.72\ppm0.22 & \textbf{73.57\ppm0.11} & 80.29\ppm0.04 \\ 
        Layer 2 & 67.48\ppm0.09 & 68.96\ppm0.02 & 78.46\ppm0.10  \\ 
        Layer 3 & 66.57\ppm0.06 & 67.97\ppm0.22 & 78.84\ppm0.17  \\ 
        Layer 4 & 65.75\ppm0.12 & 68.10\ppm0.31 & 79.37\ppm0.11   \\ 
        Layers 1-2 & \textbf{71.36\ppm0.03} & 72.93\ppm0.34 & \textbf{80.94\ppm0.13}   \\ 
        Layers 2-3 & 66.74\ppm0.24 & 68.76\ppm0.07 & 78.21\ppm0.12   \\ 
        Layers 3-4 & 65.21\ppm0.32 & 67.09\ppm0.15 &  78.34\ppm0.18   \\
        Layers 1-2-3 & 67.44\ppm0.11 & 68.59\ppm0.14 & 79.27\ppm0.05   \\
        Layers 1-2-3-4 & 68.71\ppm0.18 & 71.39\ppm0.12 & 78.38\ppm0.14   \\
    \bottomrule
    \end{tabular}
    \end{small}
    \caption{Accuracy (\%) with different combinations of projectors on 3 corruptions of CIFAR-10-C dataset. Layer $l$ means that we only use the projector after layer $l$, and Layer $l$-$l$ means that we use the sum of the two projectors' losses of these layers as total IM loss. The extractor ends at the last named layer.}
	\label{tab:LayersComparison}
\end{table}

\mypar{On the number of clusters.} As explained in Section~\ref{subseq:method}, the proxy task consists of a projector-based clustering head made by a linear mapping (implemented with a 1$\times$1 convolution) followed by a K-way softmax that projects features to a cluster probability map $\zz\! \in\![0,1]^{BWH\times K}$. In Table~\ref{tab:ClustersComparison}, we experiment with different number of clusters.
Results show that having a greater number of clusters, e.g., $K\!=\!100$, can provide a better accuracy. We also notice that having $K\!=\!10$ (corresponding to the number of classes in CIFAR-10-C) results in a competitive performance compared to other larger values, such as $K\!=\!20$ or $K\!=\!50$. This becomes a sensible approach, as projectors can help learn better class boundaries inside features.
In the next experiments, we keep $K\!=\!10$ for an efficient trade-off between performance and computational cost.

\begin{table}[!t]
    \centering
    \begin{footnotesize}
    \dorowcolors
    \begin{tabular}{l|ccccc}
    \toprule
        ~ & Gaussian & Shot & Snow & Avg$^*$ \\ \midrule
        $K$=2 & 71.58\ppm{0.12} &  73.41\ppm{0.09} & 82.98\ppm{0.10} & 80.39 \\ 
        $K$=5 & 71.10\ppm{0.09} & 72.89\ppm{0.15} & \textbf{83.76\ppm{0.09}} & 80.40 \\ 
        $K$=10 & \textbf{72.96\ppm{0.13}} & \textbf{74.55\ppm{0.12}} & 83.61\ppm{0.09} & \textbf{80.94} \\ 
        $K$=20 & 70.13\ppm{0.12} & 72.35\ppm{0.10} & 83.29\ppm{0.09} & 80.10 \\ 
        $K$=50 & 71.54\ppm{0.18} & 74.15\ppm{0.07} & 83.39\ppm{0.12} & 80.70 \\ 
        $K$=100 &  68.47\ppm{0.11} & 70.82\ppm{0.11} & 82.51\ppm{0.08} & 79.77 \\ 
        \bottomrule
        \multicolumn{5}{l}{\footnotesize$^*$: Average over the 15 corruption types}\\[2pt]
    \end{tabular}
    \end{footnotesize}
    \caption{Accuracy (\%) with different number of clusters on 3 corruptions of CIFAR-10-C dataset.}
	\label{tab:ClustersComparison}
\end{table}

\mypar{On the number of projectors per layer.} In the previous experiments, only one projector per layer was used.
Here, we evaluate whether having more projectors per layer can further improve performance.
It has been found that increasing the number of projectors per layer increases accuracy compared to using a single projector per layer (Table~\ref{tab:HeadsComparison}). However, each corruption in CIFAR-10-C can be benefited differently from different configurations. On the average, using 15 projectors on layers 1 and 2 results corresponds to the best option. In the following experiments, we compare this architecture (called ClusT3-H15) to the leading Test-Time Adaptation methods.

\begin{table*}[h!]
    \centering
    \begin{small}    
    \dorowcolors
    \begin{tabular}{l|ccccc}
    \toprule
        ~ & Head\,=\,1 & Heads\,=\,5 & Heads\,=\,10 & Heads\,=\,15 & Heads\,=\,20 \\ \midrule
        Gaussian Noise & 71.40\ppm0.26 & 72.72\ppm0.08 & 75.24\ppm0.02 & 76.01\ppm0.19 & \textbf{76.04\ppm0.20} \\ 
        Shot noise & 72.79\ppm0.04 & 74.84\ppm0.14 & 76.77\ppm0.04 & 77.67\ppm0.17 & \textbf{78.00\ppm0.05} \\ 
        Impulse Noise & 65.96\ppm0.12 & 67.78\ppm0.06 & 68.62\ppm0.07 & \textbf{69.76\ppm0.15} & 68.80\ppm0.23 \\ 
        Defocus blur & 82.77\ppm0.09 & 87.83\ppm0.09 & \textbf{87.91\ppm0.14} & 87.85\ppm0.11 & 87.86\ppm0.19 \\ 
        Glass blur & 69.65\ppm0.14 & 65.85\ppm0.04 & \textbf{71.70\ppm0.12} & 71.34\ppm0.15 & 67.26\ppm0.07 \\ 
        Motion blur & 82.03\ppm0.17 & 86.58\ppm0.07 & 86.44\ppm0.03 & 86.10\ppm0.11 & \textbf{86.91\ppm0.06} \\ 
        Zoom blur & 83.88\ppm0.09 & 86.83\ppm0.06 & \textbf{87.21\ppm0.09} & 86.68\ppm0.05 & 87.57\ppm0.06 \\ 
        Snow & 80.87\ppm0.04 & 82.68\ppm0.13 & 83.41\ppm0.06 & \textbf{83.71\ppm0.09} & 83.17\ppm0.06 \\ 
        Frost & 79.04\ppm0.07 & 81.38\ppm0.14 & 83.39\ppm0.03 & \textbf{83.69\ppm0.03} & 82.45\ppm0.11 \\ 
        Fog & 76.32\ppm0.09 & 84.40\ppm0.05 & 84.47\ppm0.14 & \textbf{85.12\ppm0.13} & 83.98\ppm0.04 \\ 
        Brightness & 89.16\ppm0.10 & \textbf{92.29\ppm0.11} & 91.91\ppm0.03 & 91.52\ppm0.02 & 91.81\ppm0.02 \\ 
        Contrast &  74.57\ppm0.25 & 85.28\ppm0.09 & 84.37\ppm0.07 & 84.40\ppm0.11 & \textbf{85.67\ppm0.08} \\ 
        Elastic transform & 80.16\ppm0.16 & 80.07\ppm0.13 & \textbf{82.33\ppm0.04} & 82.04\ppm0.17 & 82.02\ppm0.09 \\ 
        Pixelate & 80.09\ppm0.02 & 79.94\ppm0.04 & \textbf{82.75\ppm0.06} & 82.03\ppm0.09 & 82.00\ppm0.07 \\ 
        JPEG compression & 80.90\ppm0.01 & 79.86\ppm0.08 & 83.01\ppm0.08 & \textbf{83.24\ppm0.10} & 82.38\ppm0.07 \\ \midrule
        Average & 77.97 & 80.56 & 81.97 & \textbf{82.08} & 81.73 \\ 
        \bottomrule
    \end{tabular}
    \end{small}
    \caption{Accuracy (\%) with different number of projectors per layer on Layer 1 and 2 with $K\!=\!10$ on the CIFAR-10-C dataset.}
	\label{tab:HeadsComparison}
\end{table*}

\begin{figure*}[t!]
   \centering
   \includegraphics[scale=0.55]{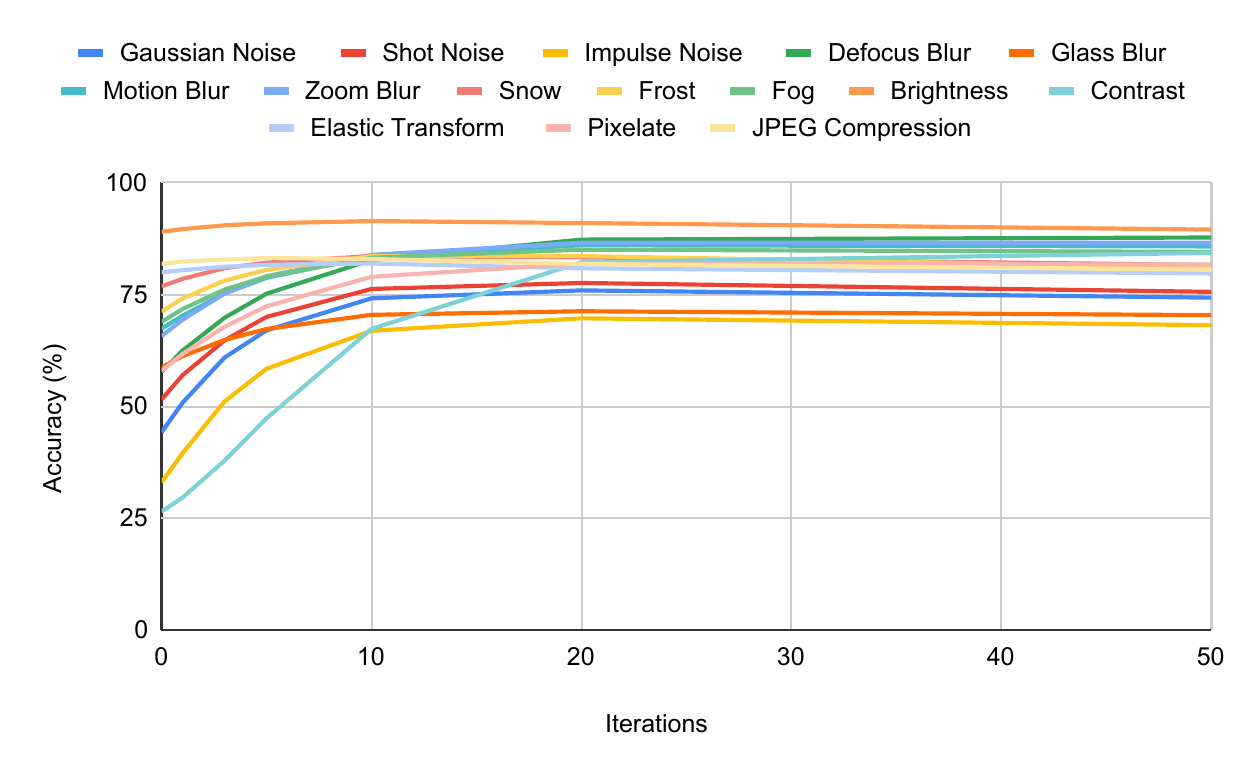}
   \caption{Evolution of accuracy for all corruptions in CIFAR-10-C.}
   \label{fig:cifar10cbar}
\end{figure*}

\mypar{Comparison of the number of iterations.} As shown in Fig~\ref{fig:cifar10cbar}, in most cases, the best accuracy is obtained after 10 or 20 iterations, depending on the corruption. Most importantly, accuracy remains constant even after 20 iterations. Furthermore, we observe that adaptation to strong corruptions (e.g., contrast) can also be done at a fast rate.

\mypar{Comparison with main TTA methods.} Several \emph{state-of-the-art} TTA/TTT techniques were chosen for comparison: TTA methods include TENT\cite{tent2021}, LAME\cite{lame2022}, and PTBN\cite{PTBN}. TTT\cite{ttt} and TTT++\cite{ttt++} are chosen for Test-Time Training. As shown in Table~\ref{tab:Cifar10cComparison}, the overall performance of ClusT3-H15 on all the corruptions outperforms ResNet50 with a gain of 28.26\% as well as all the different TTA methods.
Moreover, there is a considerably large improvement on all the individual corruptions with respect to the same baseline. A significant increase in accuracy can also be observed in most corruptions compared to previous methods, with some exceptions (e.g., Defocus blur against TTT++\cite{ttt++} or Contrast against TTT\cite{ttt}). It is however important to mention that ClusT3 differs from previous TTT methods whose self-supervised secondary task requires a higher computational overhead. TTT++, which improves considerably with respect to its predecessor TTT on Level 5, also requires preserving a queue of source feature maps to compare statistics at test-time. In comparison, ClusT3 is self-sufficient and less costly in both computation and memory. A more detailed comparison on all the corruption levels of CIFAR-10-C can be found in the supplementary material.

Table~\ref{tab:Cifar100} shows the overall performance of ClusT3 on CIFAR-100-C, in an effort to demonstrate the scalable capabilities of the method on a larger set of classes. ClusT3 mitigates the natural degradation of the ResNet50 baseline, while also outperforming \emph{state-of-the-art} methods by an important margin.


\begin{table*}[h!]
    \centering
    \begin{small}
    \dorowcolors
\resizebox{\columnwidth}{!}{
    \begin{tabular}{l|ccccccc}
    \toprule
         ~ & ResNet50  & LAME & PTBN  & TENT & TTT & TTT++ & ClusT3-H15 \\
         ~ &  & \cite{lame2022} & \cite{PTBN}  & \cite{tent2021} & \cite{ttt} & \cite{ttt++} &  \\ \midrule
        Gaussian Noise & 21.01 & 22.90 & 57.23\ppm0.13 & 57.15\ppm0.19 & 66.14\ppm0.12 & 75.87\ppm5.05 & \textbf{76.01\ppm0.19} \\ 
        Shot noise & 25.77 & 27.24 & 61.18\ppm0.03 & 61.08\ppm0.18 & 68.93\ppm0.06 & 77.18\ppm1.36 & \textbf{77.67\ppm0.17} \\ 
        Impulse Noise & 14.02 & 30.99 & 54.74\ppm0.13 & 54.63\ppm0.15 & 56.65\ppm0.03 & \textbf{70.47\ppm2.18}  & 69.76\ppm0.15 \\ 
        Defocus blur & 51.59 & 45.38 & 81.61\ppm0.07 & 81.39\ppm0.22 & \textbf{88.11\ppm0.08} & 86.02\ppm1.35 & 87.85\ppm0.11 \\ 
        Glass blur & 47.96 & 36.66 & 53.43\ppm0.11 & 53.36\ppm0.14 & 60.67\ppm0.06 & 69.98\ppm1.62 & \textbf{71.34\ppm0.15} \\ 
        Motion blur & 62.30 & 55.29 & 78.20\ppm0.28 & 78.04\ppm0.17 & 83.52\ppm0.03 & 85.93\ppm0.24 & \textbf{86.10\ppm0.11} \\ 
        Zoom blur & 59.49 & 51.40 & 80.29\ppm0.13 & 80.26\ppm0.22 & 87.25\ppm0.03 & \textbf{88.88\ppm0.95} & 86.68\ppm0.05 \\ 
        Snow & 75.41 & 66.17 & 71.59\ppm0.21 & 71.59\ppm0.04 & 79.29\ppm0.05 & 82.24\ppm1.69 & \textbf{83.71\ppm0.09} \\ 
        Frost & 63.14 & 49.98 & 68.77\ppm0.25 & 68.52\ppm0.20 & 79.84\ppm0.11 & 82.74\ppm1.63 & \textbf{83.69\ppm0.03} \\ 
        Fog & 69.63 & 64.49 & 75.79\ppm0.05 & 75.73\ppm0.10 & 84.46\ppm0.09 & 84.16\ppm0.28 & \textbf{85.12\ppm0.13} \\ 
        Brightness & 90.53 & 84.26 & 84.97\ppm0.05 & 84.77\ppm0.13 & 91.23\ppm0.08 & 89.07\ppm1.20 & \textbf{91.52\ppm0.02} \\ 
        Contrast & 33.88 & 31.50 & 80.81\ppm0.15 & 80.70\ppm0.15 & \textbf{88.58\ppm0.09} & 86.60\ppm1.39 & 84.40\ppm0.11 \\ 
        Elastic transform & 74.51 & 64.16 & 67.14\ppm0.17 & 67.13\ppm0.10 & 75.69\ppm0.10 & 78.46\ppm1.83 & \textbf{82.04\ppm0.17} \\ 
        Pixelate & 44.43 & 39.34 & 69.17\ppm0.31 & 68.70\ppm0.29 & 76.35\ppm0.19 & \textbf{82.53\ppm2.01} & 82.03\ppm0.09 \\ 
        JPEG compression & 73.61 & 66.05 & 65.86\ppm0.05 & 65.83\ppm0.07 & 73.10\ppm0.19 & 81.76\ppm1.58 & \textbf{83.24\ppm0.10} \\ \midrule
        Average & 53.82 & 49.05 & 70.05 & 69.93 & 77.32 & 81.46 & \textbf{82.08} \\
        \bottomrule 
    \end{tabular}}
    \end{small}
    \caption{Accuracy (\%) on CIFAR-10-C dataset with Level 5 corruption for ClusT3-15 compared to ResNet50, LAME, PTBN, TENT, TTT, and TTT++.}
	\label{tab:Cifar10cComparison}
\end{table*}

\begin{table}[t!]
\begin{minipage}[t]{0.38\linewidth}
\centering
\begin{footnotesize}
\dorowcolors
\setlength{\tabcolsep}{3pt}
\renewcommand{\arraystretch}{0.55}
\begin{tabular}{lc}
\toprule
\bf Method & \bf Acc. (\%) \\ 
\midrule
ResNet50 & 31.37 \\
LAME & 29.63 \\
PTBN & 54.53 \\
TENT & 54.48 \\
TTT & 51.43 \\  
Ours & \textbf{56.70} \\  
\bottomrule
\end{tabular}
\end{footnotesize}
\caption{Results on the CIFAR-100-C dataset.}
\label{tab:Cifar100}
\end{minipage}
\hfill
\begin{minipage}[t]{0.68\linewidth}
\centering
\begin{footnotesize}
\dorowcolors
        \begin{tabular}{lc}
        \toprule
\bf Method & \bf Accuracy (\%) \\ 
\midrule
ResNet50 & \textbf{88.45} \\
LAME~\cite{lame2022} & 82.68 \\
PTBN~\cite{PTBN} & 79.57\ppm0.47 \\
TENT~\cite{tent2021} & 79.69\ppm0.21 \\
TTT~\cite{ttt} & 86.30\ppm0.20 \\  
TTT++~\cite{ttt++} & 88.03\ppm0.17 \\ 
ClusT3-H5 (Ours) & 87.43\ppm0.02 \\ 
ClusT3-H15 (Ours) & 85.57\ppm0.11 \\  \bottomrule
        \end{tabular}   
\end{footnotesize}
\caption{Accuracy of compared methods on the CIFAR-10.1 dataset containing natural domain shift.}
\label{tab:Cifar10.1}
\end{minipage}
\end{table}

\begin{figure*}[ht!]
    \centering
    \begin{small}\setlength{\tabcolsep}{10pt}
    \begin{tabular}{cc}    
\includegraphics[width=0.35\linewidth]{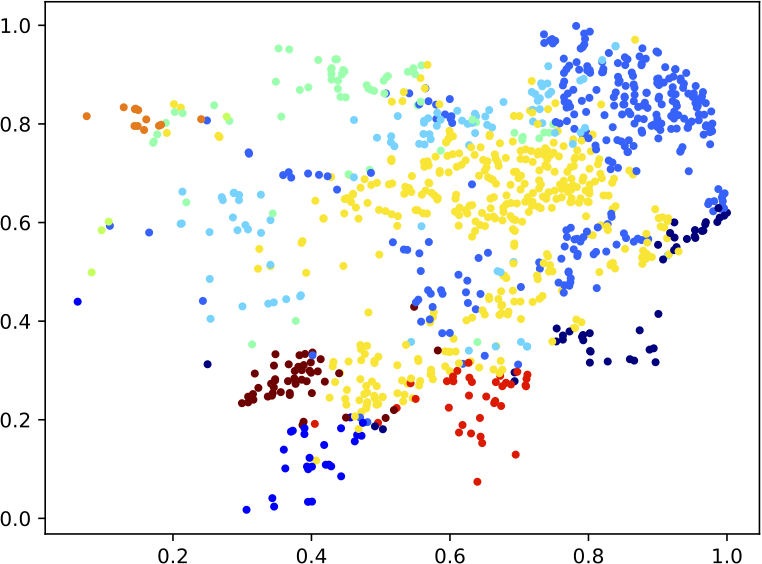} &     
    \includegraphics[width=0.35\linewidth]{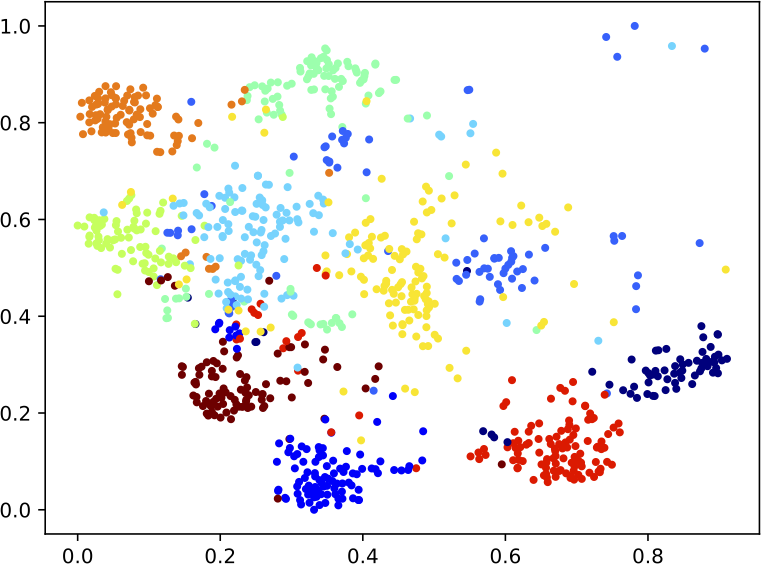}\\
    (a) Prediction (before adaptation) & (b) Prediction (after adaptation) \\[12pt]
    \includegraphics[width=0.35\linewidth]{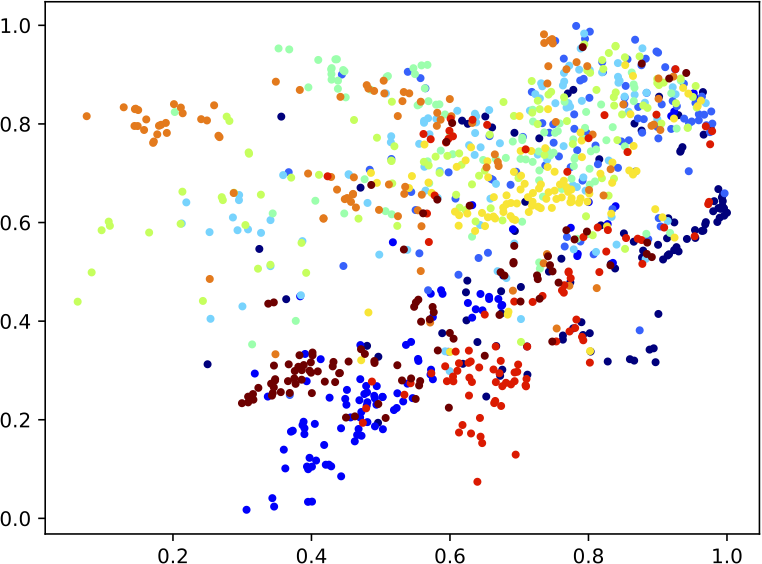} &
    \includegraphics[width=0.35\linewidth]{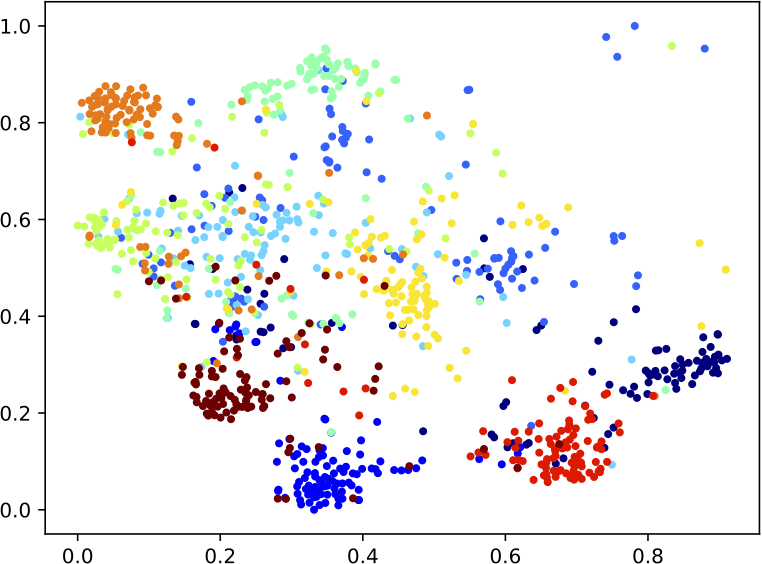}\\
    (c) Ground truth (before adaptation) & (d) Ground truth (after adaptation)\\[4pt]
    \end{tabular}
    \end{small}
    \caption{t-SNE plots of gaussian noise for the features at the output of the extractor from ClusT3 with one projector on Layer 1 and 2 each. (a) prediction of the model without adaptation. (b) prediction of the model after 20 iterations of adaptation. (c) ground truth labels without adaptation. (d) ground truth labels of adapted representations.}
    \label{fig:tSNEcifar10c}
\end{figure*}

\mypar{Visualization of adaptation.} To visualize the effect of ClusT3 during adaptation, Figure~\ref{fig:tSNEcifar10c} displays the t-SNE plots of the target feature maps before and after the adaptation with the corresponding model prediction. The projector induces the model to make better predictions by improving the clustering of the different samples' classes in the target dataset.

\subsection{Object recognition on natural domain shift}
The best configuration of ClusT3 (i.e., with 5 in that case or 15 projectors on Layer 1 and 2) is evaluated on CIFAR-10.1, which contains a more natural domain shift. A comparison is made against previous TTA methods, and as reported in Table~\ref{tab:Cifar10.1}, ClusT3 achieves a competitive accuracy despite the baseline (ResNet50) being the most accurate in this scenario. The gain of TTT++ \cite{ttt++} comes from a better pre-trained encoder thanks to the influence of contrastive learning \cite{simclr}.
This limitation can be explained by the fact that CIFAR-10 and CIFAR-10.1 are similar, thus having a smaller domain shift \cite{tttflow}.


\subsection{Object recognition on sim-to-real domain shift}

We use the VisDA-C dataset to test ClusT3 on the sim-to-real domain shift. To account for the challenge of this scenario, a slightly different projector is proposed: using two linear (1$\times$1 convolutional) layers with a ReLU activation in between. The output number of channels of the first layer is set to half the input feature maps' number of channels. This setting is named \emph{Large} projector. 
The best configuration (i.e., type of projector, number of projectors, and combination of layers) was found based on a hyperparameter study that can be found in the supplementary material. The resulting best approach consisted in using one large projector on Layer 2 (ClusT3-H1*).

\mypar{Comparison with other methods.} Our method is compared against the previously presented, popular TTT/TTA methods. For fairness, we evaluate LAME \cite{lame2022} using the three proposed affinity matrices in its original publication: LAME-L (linear affinity), LAME-K (K-NN affinity with 5 neighbors), and LAME-R (RBF affinity with 5 neighbors). As shown in Table~\ref{tab:visda}, ClusT3 achieves a higher performance than its competitors in the reproduced experiments. With respect to the baseline, ClusT3 obtains a gain of around 15.6$\%$. 

\mypar{Computational cost.} The nature of the auxiliary tasks in Test-Time Training methods can importantly impact the training efficiency. For instance, methods based on self-supervised learning might require additional forward passes, or a higher memory input, which ultimately increases the computation time. ClusT3 does not depend on additional data transformations, hence reducing execution times. We evaluate the time of one epoch of joint training (without evaluation steps) of ClusT3, utilizing 1 Large projector on top of all layers, one of the heaviest configurations. The average execution time of one epoch was $2.7947\pm0.0294$ minutes, compared to $12.4941\pm1.3994$ minutes for TTT.

\begin{table}[ht!]
\centering
\begin{small}
\dorowcolors
\begin{tabular}{cc}
\toprule
\textbf{Method}        & \textbf{Accuracy (\%)}   \\ \midrule
ResNet50      & 46.31              \\
LAME-L~\cite{lame2022}        & 22.02\ppm 0.23 \\
LAME-K~\cite{lame2022}        & 42.89\ppm 0.14 \\
LAME-R~\cite{lame2022}        & 19.33\ppm 0.11 \\
PTBN~\cite{PTBN}          & 60.33\ppm 0.04 \\
TENT~\cite{tent2021}          & 60.34\ppm 0.05 \\
TTT~\cite{ttt}           &  40.57\ppm 0.02                   \\
TTT++~\cite{ttt++}         & 60.42$^\dagger$              \\ 
ClusT3-H1* (Ours) & \textbf{61.91\ppm 0.02} \\ \bottomrule
\end{tabular}
\end{small}
\caption{Accuracy values of ClusT3 and the \emph{state-of-the-art} TTT/TTA methods on the VisDA-C dataset. $\dagger$: Result of TTT++ obtained from the original paper, were not reproducible.}
\label{tab:visda}
\end{table}

\section{Conclusion}
\label{sec:conclusion}
In this work, we proposed ClusT3, a new unsupervised Test-Time Training framework based on Information Maximization of feature latent spaces across domains. This method allows adapting the model at test-time when there is a distribution shift between the source and the target datesets. By using simple linear projectors and Mutual Information in our proxy task, we update the feature extractor to improve the accuracy at test-time.

A complete ablation study helped determine the best hyperparameters and to better understand the different possible configurations of the model. As shown in our experimental results, ClusT3 obtains a highly-competitive performance against previous TTT and TTA models. Thus, on the CIFAR-10-C dataset, ClusT3 outperforms \emph{state-of-the-art}. Surprisingly, the baseline defeats all previous methods on CIFAR-10.1, as the domain shift with respect to the source dataset is smaller and adaptation causes performance degradation. Nonetheless, ClusT3 remains competitive and robust to this scenario.

Future work includes further investigation on different architectures for the projector. As it has been shown, adding layers and nonlinearity can further improve performance in some cases. This could be due to the fact that having more complex and thus flexible projectors relaxes constraints on the feature space (e.g., balanced clusters) which can hurt the learning of a good representation for classification if too strong. Additionally, a uniform distribution has been assumed for the cluster marginal distribution. Diverging from this premise and exploring other distribution priors also constitutes an interesting line of future research. This can turn particularly useful in the scenario where adaptation to a single data sample is required.

\newpage

\section*{\centering ClusT3: Information Invariant Test-Time Training -- Suplementary Material}

\setcounter{section}{0}

\section{Results on CIFAR-10-C~\cite{cifar10c} dataset for corruption levels 1 to 4}
\label{sec:Results}

As shown in Tables~\ref{tab:Cifar10cComparisonL4}, ~\ref{tab:Cifar10cComparisonL3}, ~\ref{tab:Cifar10cComparisonL2} and ~\ref{tab:Cifar10cComparisonL1}, ClusT3 performs well on the different corruptions at different levels. It achieves a higher accuracy than ResNet50 for all corruptions, and a higher mean accuracy than all other TTA/TTT aproaches. While TTT~\cite{ttt} yields competitive performance, our method achieves a mean accuracy improvement of at least 2\% compared to this approach, on all corruption levels.

\begin{table*}[!t]
    \centering
    \begin{small}
    \dorowcolors
\resizebox{\columnwidth}{!}{
    \begin{tabular}{l|ccccccc}
    \toprule
         ~ & ResNet50  & LAME & PTBN  & TENT & TTT & TTT++ & ClusT3-H15 \\
         ~ &  & \cite{lame2022} & \cite{PTBN}  & \cite{tent2021} & \cite{ttt} & \cite{ttt++} &  \\ \midrule
        Gaussian Noise & 28.02 & 26.08 & 61.39\ppm0.10 & 61.19\ppm0.26 & 70.63\ppm0.04 & 78.70\ppm4.28 & \textbf{79.14\ppm0.03} \\ 
        Shot noise & 38.33 & 37.13 & 66.57\ppm0.06 & 66.2\ppm0.18 & 75.18\ppm0.04 & 80.12\ppm0.12 & \textbf{81.51\ppm0.15} \\ 
        Impulse Noise & 46.12 & 45.01 & 63.56\ppm0.20 & 62.98\ppm0.19 & 65.91\ppm0.04 & 70.64\ppm0.53 & \textbf{76.95\ppm0.07} \\ 
        Defocus blur & 67.33 & 67.65 & 85.48\ppm0.12 & 85.32\ppm0.18 & \textbf{91.95\ppm0.02} & 81.75\ppm0.43 & 90.33\ppm0.09 \\ 
        Glass blur & 34.42 & 32.73 & 52.26\ppm0.04 & 52.08\ppm0.15 & 60.44\ppm0.05 & 62.85\ppm0.50 & \textbf{71.09\ppm0.17} \\ 
        Motion blur & 63.71 & 64.09 & 80.78\ppm0.12 & 80.75\ppm0.09 & 86.29\ppm0.10 & 68.42\ppm1.08 & \textbf{87.87\ppm0.11} \\ 
        Zoom blur & 61.27 & 61.99 & 83.33\ppm0.11 & 83.28\ppm0.10 & \textbf{89.90\ppm0.04} & 70.74\ppm2.05 & 88.86\ppm0.04 \\ 
        Snow & 72.15 & 72.13 & 73.25\ppm0.16 & 73.17\ppm0.25 & 81.25\ppm0.02 & 52.43\ppm0.56 & \textbf{84.30\ppm0.07} \\ 
        Frost & 62.27 & 61.70 & 73.41\ppm0.22 & 73.54\ppm0.16 & 83.83\ppm0.04 & 52.80\ppm2.67 & \textbf{87.17\ppm0.07} \\ 
        Fog & 81.86 & 81.94 & 83.88\ppm0.06 & 83.81\ppm0.09 & \textbf{90.62\ppm0.05} & 41.75\ppm0.09 & 90.03\ppm0.02 \\ 
        Brightness & 87.58 & 87.71 & 86.81\ppm0.05 & 86.81\ppm0.23 & 92.87\ppm0.09 & 50.95\ppm2.19 & \textbf{92.99\ppm0.06} \\ 
        Contrast & 68.62 & 68.85 & 84.16\ppm0.09 & 84.23\ppm0.29 & 90.94\ppm0.07 & 45.28\ppm0.55 & \textbf{89.24\ppm0.07} \\ 
        Elastic transform & 67.84 & 68.25 & 76.44\ppm0.18 & 76.21\ppm0.08 & 84.03\ppm0.11 & 35.53\ppm1.51 & \textbf{86.74\ppm0.04} \\ 
        Pixelate & 56.3 & 55.83 & 76.34\ppm0.10 & 76.40\ppm0.16 & 84.92\ppm0.15 & 33.64\ppm0.83 & \textbf{87.93\ppm0.03} \\ 
        JPEG compression & 70.62 & 70.37 & 69.64\ppm0.03 & 69.54\ppm0.05 & 76.46\ppm0.04 & 28.01\ppm1.75 & \textbf{85.11\ppm0.06} \\ \midrule
        Average & 60.43 & 60.10 & 74.48 & 74.37 & 81.68 & 56.91 & \textbf{85.28} \\ \bottomrule
    \end{tabular}}
    \end{small}
    \caption{Accuracy (\%) on CIFAR-10-C dataset with Level 4 corruption for ClusT3-15 compared to ResNet50, LAME, PTBN, TENT, TTT, and TTT++.}
	\label{tab:Cifar10cComparisonL4}
\end{table*}

\begin{table*}[!t]
    \centering
    \begin{small}
    \dorowcolors
\resizebox{\columnwidth}{!}{
    \begin{tabular}{l|ccccccc}
    \toprule
         ~ & ResNet50  & LAME & PTBN  & TENT & TTT & TTT++ & ClusT3-H15 \\
         ~ &  & \cite{lame2022} & \cite{PTBN}  & \cite{tent2021} & \cite{ttt} & \cite{ttt++} &  \\ \midrule
        Gaussian Noise & 33.99 & 32.58 & 64.55\ppm0.13 & 64.67\ppm0.17 & 74.10\ppm0.09 & 80.29\ppm0.81 & \textbf{81.55\ppm0.09} \\ 
        Shot noise & 46.35 & 45.88 & 69.82\ppm0.08 & 70.04\ppm0.14 & 78.43\ppm0.07 & 82.46\ppm0.37 & \textbf{84.12\ppm0.02} \\ 
        Impulse Noise & 59.90 & 59.61 & 72.08\ppm0.14 & 71.95\ppm0.33 & 76.32\ppm0.10 & 79.20\ppm0.38 & \textbf{83.75\ppm0.01} \\ 
        Defocus blur & 79.29 & 79.58 & 87.62\ppm0.17 & 87.39\ppm0.05 & \textbf{93.25\ppm0.06} & 87.68\ppm0.38 & 91.74\ppm0.07 \\ 
        Glass blur & 47.29 & 46.44 & 63.29\ppm0.11 & 63.26\ppm0.21 & 72.09\ppm0.11 & 72.52\ppm0.56 & \textbf{79.78\ppm0.02} \\ 
        Motion blur & 63.42 & 63.72 & 81.13\ppm0.13 & 80.99\ppm0.08 & 86.48\ppm0.09 & 69.59\ppm1.38 & \textbf{88.02\ppm0.10} \\ 
        Zoom blur & 67.86 & 68.36 & 84.57\ppm0.11 & 84.34\ppm0.06 & \textbf{91.00\ppm0.02} & 73.23\ppm2.33 & 89.90\ppm0.07 \\ 
        Snow & 74.93 & 74.67 & 75.08\ppm0.14 & 75.14\ppm0.19 & 83.90\ppm0.07 & 57.96\ppm1.02 & \textbf{86.22\ppm0.07} \\ 
        Frost & 64.54 & 64.05 & 74.15\ppm0.04 & 73.98\ppm0.14 & 84.13\ppm0.10 & 49.94\ppm3.53 & \textbf{87.37\ppm0.07} \\ 
        Fog & 85.73 & 85.95 & 86.57\ppm0.09 & 86.38\ppm0.15 & \textbf{92.19\ppm0.08} & 52.89\ppm4.13 & 91.83\ppm0.01 \\ 
        Brightness & 88.93 & 88.75 & 87.50\ppm0.19 & 87.44\ppm0.01 & \textbf{93.53\ppm0.09} & 57.96\ppm1.32 & 93.31\ppm0.04 \\ 
        Contrast & 79.66 & 79.83 & 85.63\ppm0.05 & 85.46\ppm0.08 & \textbf{91.85\ppm0.09} & 53.44\ppm2.37 & 90.83\ppm0.05 \\ 
        Elastic transform & 75.67 & 75.79 & 82.72\ppm0.14 & 82.56\ppm0.15 & \textbf{90.09\ppm0.10} & 36.49\ppm3.72 & 89.33\ppm0.11 \\ 
        Pixelate & 74.83 & 75.07 & 82.17\ppm0.14 & 81.91\ppm0.13 & 89.30\ppm0.10 & 33.41\ppm3.02 & \textbf{90.23\ppm0.06} \\ 
        JPEG compression & 73.70 & 73.51 & 71.54\ppm0.09 & 71.54\ppm0.15 & 78.95\ppm0.09 & 28.82\ppm2.74 & \textbf{86.55\ppm0.06} \\ \midrule
        Average & 67.74 & 67.59 & 77.89 & 77.80 & 85.04 & 61.06 & \textbf{87.64} \\ \bottomrule
    \end{tabular}}
    \end{small}
    \caption{Accuracy (\%) on CIFAR-10-C dataset with Level 3 corruption for ClusT3-15 compared to ResNet50, LAME, PTBN, TENT, TTT, and TTT++.}
	\label{tab:Cifar10cComparisonL3}
\end{table*}

\begin{table*}[!t]
    \centering
    \begin{small}
    \dorowcolors
\resizebox{\columnwidth}{!}{
    \begin{tabular}{l|ccccccc}
    \toprule
         ~ & ResNet50  & LAME & PTBN  & TENT & TTT & TTT++ & ClusT3-H15 \\
         ~ &  & \cite{lame2022} & \cite{PTBN}  & \cite{tent2021} & \cite{ttt} & \cite{ttt++} &  \\ \midrule
        Gaussian Noise & 50.53 & 49.99 & 71.31\ppm0.16 & 71.43\ppm0.08 & 81.18\ppm0.11 & 85.41\ppm2.26 & \textbf{86.07\ppm0.08} \\ 
        Shot noise & 69.27 & 69.47 & 78.97\ppm0.19 & 79.02\ppm0.17 & 87.54\ppm0.10 & 88.79\ppm0.44 & \textbf{89.77\ppm0.04} \\ 
        Impulse Noise & 68.57 & 68.69 & 77.09\ppm0.13 & 77.03\ppm0.15 & 82.20\ppm0.13 & 84.27\ppm0.29 & \textbf{86.60\ppm0.03} \\ 
        Defocus blur & 87.45 & 87.47 & 88.20\ppm0.11 & 88.06\ppm0.06 & \textbf{93.67\ppm0.06} & 90.85\ppm0.42 & 92.87\ppm0.01 \\ 
        Glass blur & 43.26 & 42.01 & 62.66\ppm0.09 & 62.55\ppm0.11 & 71.33\ppm0.04 & 71.60\ppm1.95 & \textbf{78.81\ppm0.11} \\ 
        Motion blur & 72.98 & 73.11 & 83.51\ppm0.16 & 83.46\ppm0.10 & 89.57\ppm0.07 & 77.38\ppm1.12 & \textbf{89.78\ppm0.13} \\ 
        Zoom blur & 74.89 & 75.24 & 85.81\ppm0.21 & 85.79\ppm0.05 & \textbf{92.05\ppm0.10} & 80.30\ppm1.45 & 90.82\ppm0.04 \\ 
        Snow & 71.11 & 70.74 & 74.73\ppm0.11 & 74.69\ppm0.22 & 82.96\ppm0.08 & 68.56\ppm1.36 & \textbf{86.30\ppm0.04} \\ 
        Frost & 76.67 & 76.56 & 79.54\ppm0.15 & 79.41\ppm0.27 & 87.67\ppm0.03 & 63.66\ppm3.39 & \textbf{90.27\ppm0.10} \\ 
        Fog & 88.51 & 88.47 & 87.62\ppm0.10 & 87.60\ppm0.17 & \textbf{93.23\ppm0.04} & 64.26\ppm3.37 & 93.07\ppm0.04 \\ 
        Brightness & 89.75 & 89.57 & 88.09\ppm0.03 & 87.97\ppm0.14 & \textbf{93.69\ppm0.08} & 67.19\ppm1.23 & 93.64\ppm0.01 \\ 
        Contrast & 84.58 & 84.79 & 86.19\ppm0.17 & 86.41\ppm0.04 & \textbf{92.50\ppm0.12} & 62.90\ppm1.93 & 92.00\ppm0.01 \\ 
        Elastic transform & 82.10 & 82.26 & 83.69\ppm0.13 & 83.68\ppm0.08 & \textbf{90.98\ppm0.12} & 50.06\ppm2.37 & 90.37\ppm0.01 \\ 
        Pixelate & 81.04 & 80.94 & 82.92\ppm0.14 & 83.01\ppm0.07 & 90.61\ppm0.15 & 43.33\ppm3.31 & \textbf{91.28\ppm0.09} \\ 
        JPEG compression & 76.06 & 76.04 & 73.63\ppm0.02 & 73.56\ppm0.13 & 81.37\ppm0.11 & 28.26\ppm2.78 & \textbf{87.86\ppm0.08} \\ \midrule
        Average & 74.45 & 74.36 & 80.26 & 80.24 & 87.37 & 68.45 & \textbf{89.30} \\ \bottomrule
    \end{tabular}}
    \end{small}
    \caption{Accuracy (\%) on CIFAR-10-C dataset with Level 2 corruption for ClusT3-15 compared to ResNet50, LAME, PTBN, TENT, TTT, and TTT++.}
	\label{tab:Cifar10cComparisonL2}
\end{table*}

\begin{table*}[!t]
    \centering
    \begin{small}
    \dorowcolors
\resizebox{\columnwidth}{!}{
    \begin{tabular}{l|ccccccc}
    \toprule
         ~ & ResNet50  & LAME & PTBN  & TENT & TTT & TTT++ & ClusT3-H15 \\
         ~ &  & \cite{lame2022} & \cite{PTBN}  & \cite{tent2021} & \cite{ttt} & \cite{ttt++} &  \\ \midrule
        Gaussian Noise & 71.38 & 71.54 & 79.22\ppm0.13 & 79.52\ppm0.12 & 88.38\ppm0.12 & 90.14\ppm1.05 & \textbf{90.35\ppm0.05} \\ 
        Shot noise & 80.39 & 80.44 & 82.21\ppm0.05 & 82.18\ppm0.15 & 90.43\ppm0.02 & 90.89\ppm0.29 & \textbf{91.42\ppm0.02} \\ 
        Impulse Noise & 80.04 & 80.05 & 82.39\ppm0.08 & 82.48\ppm0.15 & 88.23\ppm0.02 & 87.76\ppm0.06 & \textbf{90.51\ppm0.06} \\ 
        Defocus blur & 90.17 & 89.96 & 88.28\ppm0.04 & 88.26\ppm0.15 & \textbf{93.89\ppm0.04} & 91.51\ppm0.48 & 93.72\ppm0.09 \\ 
        Glass blur & 40.96 & 39.79 & 63.19\ppm0.05 & 63.22\ppm0.15 & 71.12\ppm0.07 & 72.12\ppm2.13 & \textbf{790.1\ppm0.21} \\ 
        Motion blur & 82.78 & 82.75 & 85.99\ppm0.09 & 85.89\ppm0.08 & \textbf{91.97\ppm0.05} & 84.11\ppm0.91 & 91.50\ppm0.13 \\ 
        Zoom blur & 78.58 & 78.90 & 86.19\ppm0.06 & 86.23\ppm0.04 & \textbf{92.21\ppm0.08} & 81.76\ppm1.38 & 90.87\ppm0.04 \\ 
        Snow & 83.45 & 83.33 & 82.94\ppm0.13 & 82.84\ppm0.35 & 88.90\ppm0.04 & 75.89\ppm0.75 & \textbf{90.33\ppm0.02} \\ 
        Frost & 84.84 & 84.48 & 83.88\ppm0.15 & 83.71\ppm0.24 & 91.17\ppm0.03 & 71.54\ppm3.13 & \textbf{92.19\ppm0.06} \\ 
        Fog & 90.15 & 90.10 & 88.31\ppm0.13 & 88.05\ppm0.06 & \textbf{93.71\ppm0.09} & 70.58\ppm1.29 & 93.64\ppm0.01 \\ 
        Brightness & 90.35 & 90.19 & 88.28\ppm0.09 & 88.35\ppm0.25 & \textbf{93.90\ppm0.06} & 64.40\ppm2.69 & 93.83\ppm0.05 \\ 
        Contrast & 89.52 & 89.33 & 87.98\ppm0.09 & 87.93\ppm0.08 & \textbf{93.61\ppm0.05} & 53.60\ppm3.80 & \textbf{93.61\ppm0.03} \\ 
        Elastic transform & 82.46 & 82.57 & 83.29\ppm0.17 & 83.28\ppm0.27 & \textbf{90.55\ppm0.09} & 39.92\ppm1.52 & 90.33\ppm0.06 \\ 
        Pixelate & 87.27 & 87.15 & 85.79\ppm0.12 & 85.81\ppm0.17 & 92.24\ppm0.01 & 36.04\ppm3.47 & \textbf{92.74\ppm0.04} \\ 
        JPEG compression & 82.03 & 81.73 & 79.72\ppm0.10 & 79.82\ppm0.14 & 86.86\ppm0.08 & 30.90\ppm1.18 & \textbf{90.90\ppm0.01} \\ \midrule
        Average & 80.96 & 80.82 & 83.17 & 83.17 & 89.81 & 69.41 & \textbf{91.00} \\ \bottomrule
    \end{tabular}}
    \end{small}
    \caption{Accuracy (\%) on CIFAR-10-C dataset with Level 1 corruption for ClusT3-15 compared to ResNet50, LAME, PTBN, TENT, TTT, and TTT++.}
	\label{tab:Cifar10cComparisonL1}
\end{table*}

\section{Hyperparameters search on VisDA-C}

We perform the hyperparameter search to find an efficient configuration for VisDA-C. We evaluate to up to 20 iterations, using all the different individual layers, as well as combinations of them. Specifically, we tested the following settings: 
\begin{itemize}
    \item A single normal projector (one $1\times1$ convolution) in Table~\ref{tab:visda1projsmall};
    \item Five normal projectors in Table~\ref{tab:visda5projsmall};
    \item Ten normal projectors in Table~\ref{tab:visda10projsmall};
    \item A single Large projector (two 1 $1\times1$ 1 convolutions with ReLU in between) in Table~\ref{tab:visda1projlarge};
    \item Five Large projectors in Table~\ref{tab:visda5projlarge}.
\end{itemize}
As observed in these results, our ClusT3 method obtains significant improvements in different settings. For this dataset, the best accuracy is achieved using a single large projector applied to the second layer.

\begin{table*}[t!]
\centering
\dorowcolors
\begin{small}
\begin{tabular}{c|cccccccc}
\hline
\multicolumn{1}{l|}{}  & \multicolumn{8}{c}{\textbf{Layers}}                                                                              \\ \hline
\textbf{Iterations}    & \textbf{1} & \textbf{2} & \textbf{3} & \textbf{4} & \textbf{1, 2} & \textbf{2, 3} & \textbf{3, 4} & \textbf{All} \\ \hline
\textbf{No adaptation} & 45.31      & 45.57      & 45.67      & 47.09      & 45.66         & 44.27         & 38.17         & 42.89        \\
\textbf{1}             & 49.07      & 48.73      & 48.96      & 52.96      & 50.13         & 47.85         & 49.35         & 48.16        \\
\textbf{3}             & 52.53      & 54.31      & 53.99      & 56.23      & 52.18         & 54.7          & 55.57         & 54.23        \\
\textbf{10}            & 57.67      & 58.19      & 58.27      & 58.79      & 57.78        & 58.74         & 58.31         & 57.56        \\
\textbf{20}            & 56.84      & \textbf{59.82}      & 56.61      & 57.34      & 57.41         & 57.63         & 56.31         & 55.87        \\ \hline
\end{tabular}
\end{small}
\caption{Accuracy ($\%$) values on VisDA-C with 1 normal projector on different layers.}
\label{tab:visda1projsmall}
\end{table*}

\begin{table*}[t!]
\centering
\dorowcolors
\begin{small}
\begin{tabular}{c|cccccccc}
\hline
\multicolumn{1}{l|}{}  & \multicolumn{8}{c}{\textbf{Layers}}                                                                              \\ \hline
\textbf{Iterations}    & \textbf{1} & \textbf{2} & \textbf{3} & \textbf{4} & \textbf{1, 2} & \textbf{2, 3} & \textbf{3, 4} & \textbf{All} \\ \hline
\textbf{No adaptation} & 45.31      & 45.57      & 45.67      & 47.09      & 45.66         & 44.27         & 38.17         & 42.89        \\
\textbf{1}             & 49.57      & 49.84      & 50.28      & 50.69      & 49.58         & 47.38         & 41.39         & 47.72        \\
\textbf{3}             & 54.39      & 54.51      & 55.73      & 54.47      & 53.99         & 52.11         & 47.15         & 53.75        \\
\textbf{10}            & 58.77      & 58.85      & 60.28      & 58.31      & 58.47         & 57.34         & 55.94         & 57.89       \\
\textbf{20}            & 57.66      & \textbf{61.02}      & 58.43      & 57.17      & 56.83         & 56.52         & 56.31         & 55.87        \\ \hline
\end{tabular}
\end{small}
\caption{Accuracy ($\%$) values on VisDA-C with 5 normal projectors on different layers.}
\label{tab:visda5projsmall}
\end{table*}

\begin{table*}[t!]
\centering
\dorowcolors
\begin{small}
\begin{tabular}{c|cccccccc}
\hline
\multicolumn{1}{l|}{}  & \multicolumn{8}{c}{\textbf{Layers}}                                                                              \\ \hline
\textbf{Iterations}    & \textbf{1} & \textbf{2} & \textbf{3} & \textbf{4} & \textbf{1, 2} & \textbf{2, 3} & \textbf{3, 4} & \textbf{All} \\ \hline
\textbf{No adaptation} & 46.51      & 46.02      & 44.69      & 45.97      & 44.22         & 46.02         & 43.2          & 41.05        \\
\textbf{1}             & 49.28      & 50.31      & 49.38      & 50.38      & 48.43         & 50.35         & 46.19         & 44.25        \\
\textbf{3}             & 54.79      & 55.69      & 55.31      & 55.19      & 53.34         & 55.71         & 51.35         & 44.19        \\
\textbf{10}            & 61.13      & 60.93      & 60.61      & 59.6       & 59.62         & 60.78         & 59.09         & 59.06        \\
\textbf{15}            & \textbf{61.53}      & 61.16      & 60.59      & 59.97      & 60.37         & 60.55         & 59.82         & 59.97        \\
\textbf{20}            & 61.33      & 60.86      & 59.92      & 59.8       & 60.22         & 59.89         & 59.73         & 58.98        \\ \hline
\end{tabular}
\end{small}
\caption{Accuracy ($\%$) values on VisDA-C with 10 normal projectors on different layers.}
\label{tab:visda10projsmall}
\end{table*}

\begin{table*}[t!]
\centering
\dorowcolors
\begin{small}
\begin{tabular}{c|cccccccc}
\hline
\multicolumn{1}{l|}{}  & \multicolumn{8}{c}{\textbf{Layers}}                                                                              \\ \hline
\textbf{Iterations}    & \textbf{1} & \textbf{2} & \textbf{3} & \textbf{4} & \textbf{1, 2} & \textbf{2, 3} & \textbf{3, 4} & \textbf{All} \\ \hline
\textbf{No adaptation} & 43.91      & 46.41      & 46         & 42.46      & 46.54         & 45.45         & 44.27         & 44.09        \\
\textbf{1}             & 48.28      & 51.79      & 49.82      & 45.96      & 51.06         & 49.23         & 49.41         & 49.36        \\
\textbf{3}             & 54.23      & 56.72      & 54.24      & 51.62      & 56.6          & 54.36         & 54.89         & 49.43        \\
\textbf{10}            & 60.04      & 61.72      & 59.16      & 59.56      & 60.93         & 60.64         & 61.12         & 60.18        \\
\textbf{15}            & 60.25      & \textbf{61.93}      & 59.44      & 60.16      & 60.81         & 60.91         & 61.64         & 60.27        \\
\textbf{20}            & 59.98      & 61.57      & 59.14      & 59.88      & 60.31         & 60.72         & 61.16         & 59.92        \\ \hline
\end{tabular}
\end{small}
\caption{Accuracy ($\%$) values on VisDA-C with 1 Large projector on different layers.}
\label{tab:visda1projlarge}
\end{table*}

\begin{table*}[t!]
\centering
\dorowcolors
\begin{small}
\begin{tabular}{c|cccccccc}
\hline
\multicolumn{1}{l|}{}  & \multicolumn{8}{c}{\textbf{Layers}}                                                                                                        \\ \hline
\textbf{Iterations}    & \textbf{1} & \textbf{2} & \textbf{3} & \textbf{4} & \textbf{1, 2} & \textbf{2, 3} & \textbf{3, 4} & \textbf{All}                           \\ \hline
\textbf{No adaptation} & 46.57     & 44.66     & 46.01     & 43.86     & 45.21        & 46.37        & 46.58        & 46.81                                 \\
\textbf{1}             & 50.67     & 47.68     & 50.16     & 48.61     & 49.70         & 49.46        & 49.46        & 52.34                                 \\
\textbf{3}             & 56.18     & 52.77     & 54.90      & 54.27     & 54.65        & 52.88        & 53.43        & 52.29                                 \\
\textbf{10}            & 61.45     & 61.03     & 59.87     & 60.51     & 59.57        & 59.95        & 58.12        & \textbf{61.69} \\
\textbf{15}            & 61.48     & 61.54     & 60.40      & 60.96     & 60.16        & 61.44        & 58.83        & 61.51                                 \\
\textbf{20}            & 60.91     & 60.90      & 59.81     & 60.49     & 59.83        & 59.30         & 58.69        & 60.66                                 \\ \hline
\end{tabular}
\end{small}
\caption{Accuracy ($\%$) values on VisDA-C with 5 Large projectors on different layers.}
\label{tab:visda5projlarge}
\end{table*}

\bibliographystyle{unsrtnat}
\bibliography{main}  






\end{document}